\documentclass[letterpaper, 10 pt, conference]{ieeeconf}  

\IEEEoverridecommandlockouts                              

\usepackage{amsmath,amsfonts}
\usepackage{amssymb}
\usepackage{graphicx}
\usepackage{booktabs}
\usepackage{multirow}
\usepackage{tabularx}
\usepackage{makecell}
\usepackage{cite}

\usepackage[export]{adjustbox}
\usepackage{tikz}            %

\usepackage{hyperref}

\title{\LARGE \bf
CONTACT: CONtact-aware TACTile Learning for Robotic Disassembly
}

\author{
Yosuke Saka$^{*}$, 
Jyun-Chi Hu$^{*}$, 
Adeesh Desai$^{*}$, 
Zhiyuan Zhang$^{*}$,
Bihao Zhang,
Quan Khanh Luu,\\
Md Rakibul Islam Prince,
Minghui Zheng, 
Yu She$^{\dagger}$%
\thanks{This work was partially supported by the United States Department of Agriculture (USDA) under Grant Nos. 2023-67021-39072 and 2024-67021-42878, and by the National Science Foundation (NSF) under Grant Nos. 2423068 and 2520136.}
\thanks{$^{*}$These authors contributed equally. 
$^{\dagger}$Corresponding author.}%
\thanks{Yosuke, Jyun-Chi, Adeesh, Zhiyuan, Quan, Prince, and Yu are with the Department of Industrial Engineering, Purdue University, West Lafayette, IN 47907, USA 
{\tt\footnotesize \{ysaka, hu1148, desai274, zhan5570, luu15, prince26, shey\}@purdue.edu}}%
\thanks{Bihao and Minghui are with the Department of Mechanical Engineering, Texas A\&M University, College Station, TX 77843, USA 
{\tt\footnotesize \{bhzhang, mhzheng\}@tamu.edu}}%
}

\begin{document}

\maketitle
\thispagestyle{empty}
\pagestyle{empty}

\begin{abstract}
Robotic disassembly involves contact-rich interactions in which successful manipulation depends not only on geometric alignment but also on force-dependent state transitions. While vision-based policies perform well in structured settings, their reliability often degrades in tight-tolerance, contact-dominated, or deformable scenarios. In this work, we systematically investigate the role of tactile sensing in robotic disassembly through both simulation and real-world experiments.
We construct five rigid-body disassembly tasks in simulation with increasing geometric constraints and extraction difficulty. We further design five real-world tasks, including three rigid and two deformable scenarios, to evaluate contact-dependent manipulation.
Within a unified learning framework, we compare three sensing configurations: Vision Only, Vision + tactile RGB (TacRGB), and Vision + tactile force field (TacFF).
Across both simulation and real-world experiments, TacFF-based policies consistently achieve the highest success rates, with particularly notable gains in contact-dependent and deformable settings.
Notably, naive fusion of TacRGB and TacFF underperforms either modality alone, indicating that simple concatenation can dilute task-relevant force information. Our results show that tactile sensing plays a critical, task-dependent role in robotic disassembly, with structured force-field representations being particularly effective in contact-dominated scenarios.
\textbf{Project Website:} \url{https://vict0rhu.github.io/CONTACT-Website/}
\end{abstract}

\section{INTRODUCTION}
Robotic disassembly is a critical capability for recycling, maintenance, and remanufacturing.
Unlike many pick-and-place tasks that primarily rely on free-space motion and geometric positioning, disassembly often requires breaking existing contact constraints, overcoming frictional resistance, and disengaging interlocking geometries.
Successful execution therefore depends not only on geometric alignment but also on detecting force-dependent state transitions.
These characteristics make disassembly fundamentally sensitive to contact transitions and force modulation.
Classical model-based approaches have achieved success in structured insertion tasks such as peg-in-hole assembly~\cite{Whitney_1982, park2013intuitive}, but rely on accurate modeling of geometry and contact conditions.
In real-world disassembly scenarios, objects exhibit significant variations in geometry, wear conditions, and material properties~\cite{foo2022challenges}, limiting the robustness of purely analytical methods.

Recent advances in learning-based policies have demonstrated strong performance in structured manipulation tasks~\cite{zhao2023learning, diffusion_policy, DP3, zhang2025canonical, equiform, zhang2025vibecheck, kim2024openvla, intelligence2025pi_}.
Among these approaches, reinforcement learning (RL) has been applied to force-sensitive disassembly behaviors such as twist-pull operations~\cite{zang2025augmenting}, enabling robots to learn complex contact interactions through trial and error~\cite{zhang2025safe}.
However, RL-based methods are often sensitive to contact dynamics and reward design, and typically require extensive domain randomization or shaping to generalize across variations in geometry and resistance, which are common in real-world disassembly scenarios~\cite{tsuji2025survey}.
Diffusion-based policies model action generation as a denoising process conditioned on observations, enabling flexible multi-modal conditioning~\cite{diffusion_policy, DP3}. These approaches have recently been extended to disassembly tasks, including compliant object prying and language-guided manipulation~\cite{kang2025roboticcompliantobjectprying, kang2025task}. 

\begin{figure}[t]
\centering
\includegraphics[width=1.0\linewidth]{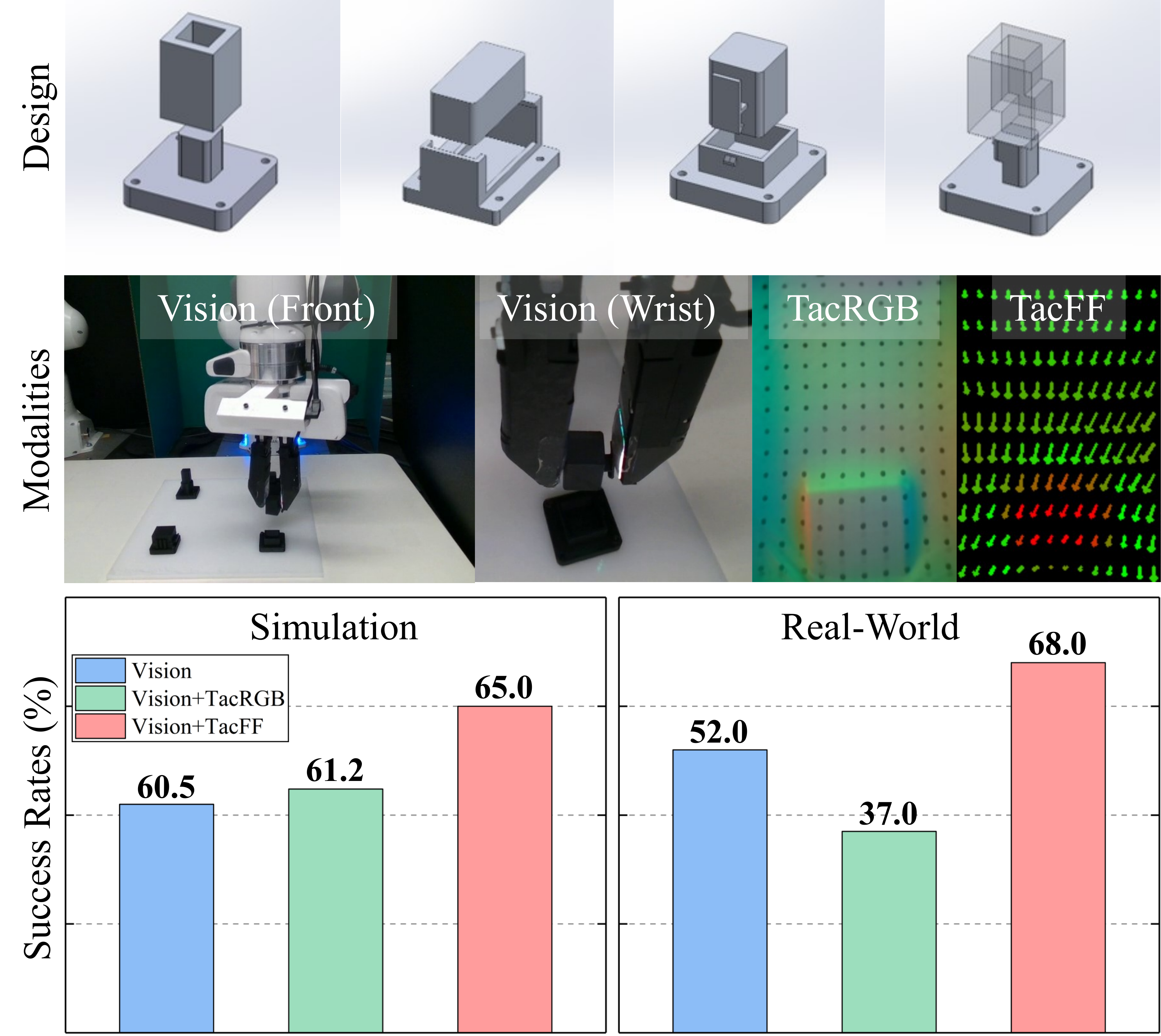} 
\caption{Structured task design and multimodal evaluation for robotic disassembly. Representative tasks span increasing contact complexity. We compare three sensing configurations: Vision Only, Vision + tactile RGB images (TacRGB), and Vision + tactile force-field representations (TacFF). Across both simulation and real-world experiments, TacFF achieves the highest overall success rates, highlighting the importance of structured force encoding in contact-rich and deformable disassembly.
}
\label{fig:abstract}
\end{figure}

\begin{figure*}[t]
    \centering
    \includegraphics[width=1.0\linewidth]{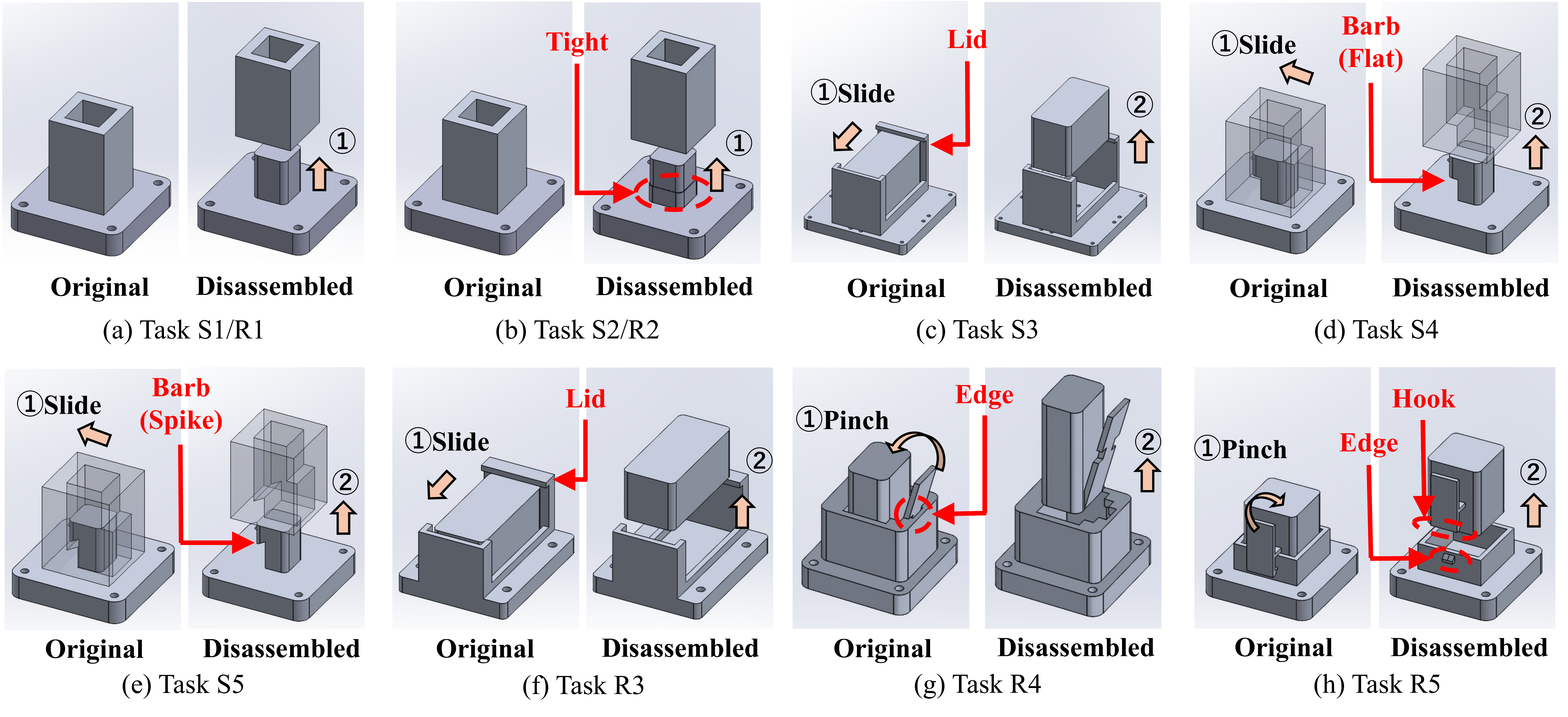}
    \caption{
    Object geometries and interaction primitives of simulation and real-world disassembly tasks. Each task consists of an initial (Original) and goal (Disassembled) configuration. S1/R1 and S2/R2 share identical shapes. S3/R3 are similar but exhibit minor geometric differences (S3 is larger than R3). S4–S5 (simulation) and R4–R5 (real-world) represent distinct tasks without direct correspondence. The tasks involve diverse interaction modes including pulling, sliding, and pinching disengagement.
    }
    \label{fig:objects}
\end{figure*}

Despite these advances, most existing systems rely predominantly on vision. While high-capacity visual encoders~\cite{zou2023segment, oquab2023dinov2, simeoni2025dinov3, wu2024point} and large-scale demonstration datasets~\cite{khazatsky2024droid, chen2025robotwin} often make vision-only policies appear sufficient, visual observations alone may fail to capture subtle force interactions and contact transitions in contact-rich scenarios~\cite{luu2025manifeel, schneider2025tactile, hu2025vibrissae}. In disassembly tasks where resistance changes, tight tolerances, or deformable releases are present, these unobserved contact dynamics can critically affect policy robustness.

Tactile sensing has therefore emerged as a complementary modality for contact-dominated manipulation~\cite{GelSight, GelTip, zhang2024gelroller, zhang2023gelflow, zhou2026tactile, 10242327, 10965524}. Recent work has integrated tactile feedback into learning-based policies~\cite{luu2025manifeel, higuera2026visuo}, demonstrating improved robustness in force-sensitive tasks. However, tactile observations can be represented in multiple forms, ranging from high-resolution deformation images to compact force-field encodings~\cite{zhu2025residual, heng2025vitacformer}. While these representations have shown promise in general manipulation settings, their relative advantages in structured robotic disassembly remain under-explored.

In this work, we systematically investigate the role of tactile sensing in robotic disassembly by addressing two central questions:

1. When does tactile sensing significantly improve performance over vision alone?

2. Which tactile representation is most effective for contact-rich disassembly?

To answer these questions, we design a structured evaluation framework spanning both simulation and real-world settings.
An overview of the proposed task suite and sensing modalities is shown in Fig.~\ref{fig:abstract}.
In simulation, we construct five rigid-body disassembly tasks with progressively increasing geometric constraints and contact dependencies under controlled physical assumptions.
In the real world, we introduce five additional tasks, including deformable mechanisms that require compliance-dependent release and precise force modulation. This progression, from loose extraction to tight tolerances, interlocking constraints, and deformable release, enables systematic analysis of sensing modalities across increasing contact complexity.
Within a unified policy learning framework, we evaluate three sensing configurations: Vision Only, Vision + tactile RGB images (TacRGB), and Vision + tactile force-field representations (TacFF). By training policies under identical conditions, we directly compare modality effects across tasks in both simulation and real-world experiments.

Our results reveal a clear task-dependent pattern. In geometry-dominant scenarios, vision-only policies perform competitively. As constraints tighten or deformable interactions are introduced, TacFF consistently improves robustness in both simulation and real-world experiments.
Notably, naive fusion of TacRGB and TacFF does not yield further gains, suggesting that structured force representations, rather than increased observation dimensionality, are critical for contact-dominated disassembly.

In summary, this work makes three main contributions:

1. We introduce a structured set of disassembly task suites spanning rigid and deformable interactions with progressively increasing contact complexity.

2. We provide a systematic comparison of visual and tactile representations under a unified policy framework across simulation and real-world settings.

3. We identify task-dependent modality effects, showing that compact force-field representations outperform high-dimensional tactile imagery in contact-dominated disassembly, and that naive multimodal fusion can even degrade performance.

\section{Structured Disassembly Tasks and Multimodal Policy Learning}
\begin{figure*}
    \centering
    \includegraphics[width=1.0\linewidth]{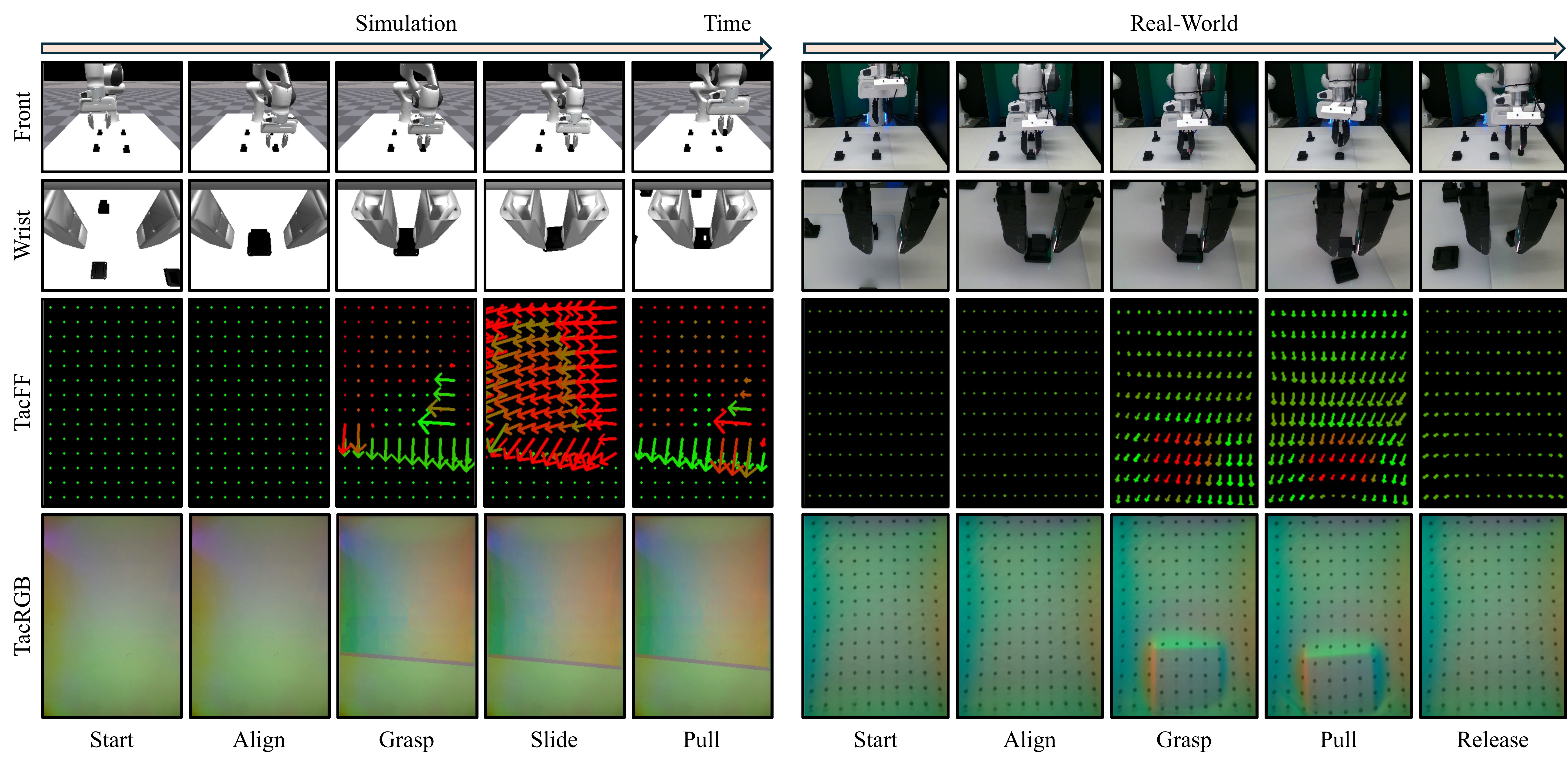}
    \caption{Comparison of multimodal observations over time in simulation (left) and real-world (right) disassembly. Columns denote task stages, and rows correspond to the front view, wrist view, TacFF, and TacRGB. TacFF visualizes distributed shear (arrow length) and normal force (color, green to red), highlighting contact-state evolution during manipulation. All modalities are recorded at 10 Hz.}
    \label{fig:time}
\end{figure*}
\subsection{Disassembly Task Suite}
Disassembly tasks inherently involve contact-rich interactions, geometric constraints, and force-dependent state transitions.
To systematically evaluate the contribution of tactile sensing under varying physical conditions, we design both simulation and real-world task sets with progressively increasing physical complexity.
The simulation tasks involve only rigid-body interactions, while the real-world tasks incorporate deformable components and compliance-induced contact dynamics beyond rigid-body modeling.
These structured simulation and real-world task suites form a unified evaluation framework spanning rigid extraction, tolerance-sensitive insertion, asymmetric resistance, multi-stage release, and deformable interactions.

\subsubsection{Simulation Tasks}
In simulation, we construct five rigid-body disassembly tasks (S1–S5) under controlled physical assumptions. Deformable geometries, material compliance, and deformation-induced contact-state transitions are not modeled. Instead, each task isolates key structural aspects of disassembly, such as extraction tolerance and staged release, within a fully observable and deterministic setting.
The simulation environment enables precise control of geometric alignment, allowing us to evaluate policy behavior in geometry-dominant scenarios without additional uncertainties from deformable objects or unmodeled contact dynamics.

The object geometries are shown in Fig.~\ref{fig:objects}(a)–(e). Although all tasks involve rigid components, they progressively increase in geometric and interaction complexity:
\begin{itemize}
    \item \textbf{S1 (Vertical Pull, Loose Socket):} A simple extraction task with generous clearance, primarily evaluating geometric alignment and grasp stability.
    \item \textbf{S2 (Vertical Pull, Tight Socket):} Reduced tolerance compared to S1, requiring more precise alignment and friction-aware force control.
    \item \textbf{S3 (Loose Plug with Lid):} A multi-stage task where the agent must first disengage a lid-like constraint before completing extraction. The object is slightly enlarged relative to its real-world counterpart to simplify grasp acquisition.
    \item \textbf{S4 (Vertical Pull, Flat Barb):} Introduces directional resistance through a flat barb structure, creating asymmetric contact forces during extraction.
    \item \textbf{S5 (Vertical Pull, Spike Barb):} Incorporates a spike-shaped barb that requires careful motion to avoid collision during extraction.
\end{itemize}
Together, S1–S5 span increasing levels of geometric constraint and resistance asymmetry while remaining within rigid-body assumptions, forming a controlled setting for evaluating modality contributions in simplified disassembly scenarios.

\subsubsection{Real-World Tasks}
To evaluate policy performance under physically richer and less controlled conditions, we design five real-world disassembly tasks (R1–R5). Compared to the simulation setting, these tasks incorporate friction variability and selectively include deformable components that introduce compliance-induced contact-state transitions.

The real-world object geometries are shown in Fig.~\ref{fig:objects}(a), (b), and (f)–(h). Tasks R1 and R2 share identical geometries with S1 and S2, respectively. Task R3 is structurally similar to S3, although the simulation version is slightly enlarged to simplify grasp acquisition. The five real-world tasks progressively increase in physical interaction complexity:
\begin{itemize}
    \item \textbf{R1 (Vertical Pull, Loose Socket):} Rigid-body extraction with sufficient clearance.
    \item \textbf{R2 (Vertical Pull, Tight Socket):} Tight extraction requiring friction-aware force control and precise alignment under realistic contact variability.
    \item \textbf{R3 (Loose Plug with Lid):} A multi-stage rigid-body task requiring sustained contact and sequential disengagement.
    \item \textbf{R4 (Vertical Push Tab):} A deformable tab that must be compressed prior to release, introducing compliance-dependent state transitions.
    \item \textbf{R5 (Vertical Clip):} An elastic clip mechanism requiring controlled deformation and localized force application to disengage a hook.
\end{itemize}
Unlike the purely rigid simulation setting, R4 and R5 introduce deformable structures in which successful manipulation depends on detecting subtle contact-state changes and force redistribution. These tasks particularly highlight the importance of tactile sensing for capturing compliance effects and incipient release events that are difficult to infer from vision alone.
Together, R1–R5 extend the controlled simulation tasks toward realistic physical variability, enabling systematic analysis of modality contributions across rigid and deformable interaction regimes.

\subsection{Multimodal Observations}
\begin{figure}
    \centering
    \includegraphics[width=1\linewidth]{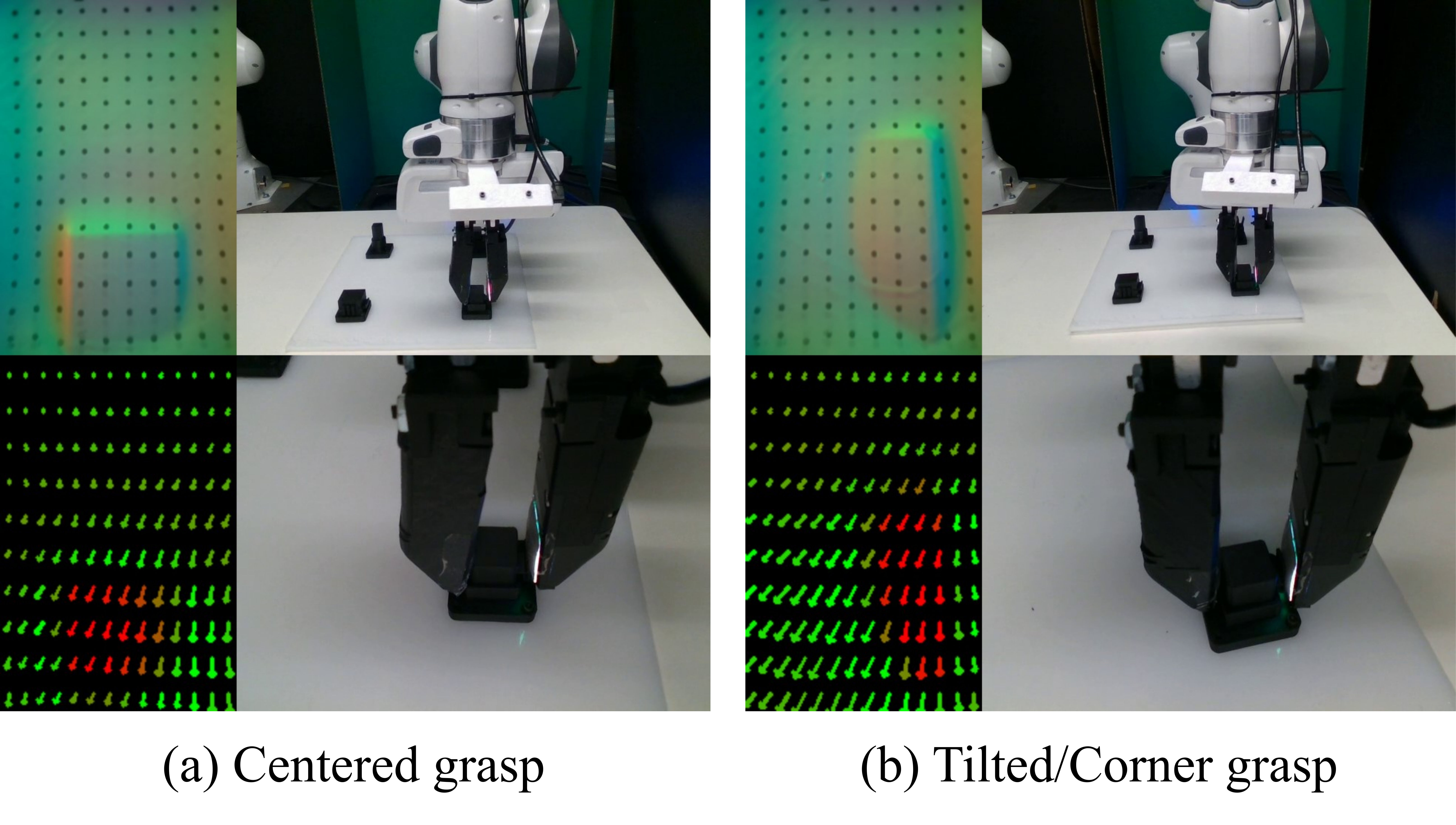}
    \caption{Visual ambiguity versus tactile disambiguation during grasp in Task R5.  Despite similar RGB appearance due to low contrast and self-occlusion, TacRGB and TacFF clearly reveal grasp quality: corner or tilted contacts produce localized indentation in TacRGB and concentrated shear patterns in TacFF.}
    \label{fig:grasp_compare}
\end{figure}

Our system integrates visual and tactile sensing to capture complementary information during disassembly. Visual input is provided by a front-view camera for global context and a wrist-mounted camera for local gripper observations. Tactile sensing is obtained from a GelSight sensor mounted on the end-effector.

We represent the tactile signal in two complementary forms. \textbf{TacRGB} preserves high-resolution surface deformation images. \textbf{TacFF} encodes a compact force-field representation computed from optical flow and depth reconstruction, consisting of spatially distributed shear components and normal force over a grid. In TacFF, arrow direction and length represent shear direction and magnitude, while color (green to red) indicates increasing normal force.

Fig.~\ref{fig:time} illustrates the temporal evolution of multimodal observations during disassembly, comparing simulation (left) and real-world execution (right). Despite different physical regimes, both exhibit consistent force-dependent tactile patterns. Prior to contact, tactile signals remain near zero. Upon grasp, localized normal forces and directional shear emerge. During sliding and pulling, shear patterns redistribute across the contact surface, reflecting evolving contact constraints. These structured force patterns directly encode contact-state transitions that are not explicitly visible in RGB images.

Despite rich visual input, grasp quality can remain ambiguous due to low contrast and self-occlusion. As shown in Fig.~\ref{fig:grasp_compare}, a centered grasp and a tilted corner grasp appear visually similar. However, tactile sensing reveals clear differences: TacRGB shows localized indentation, and TacFF exhibits asymmetric shear concentration in the tilted case. This demonstrates the importance of tactile feedback for resolving contact conditions in disassembly.

\subsection{Vision–Tactile Policy Learning}
\begin{figure}
    \centering
    \includegraphics[width=1\linewidth]{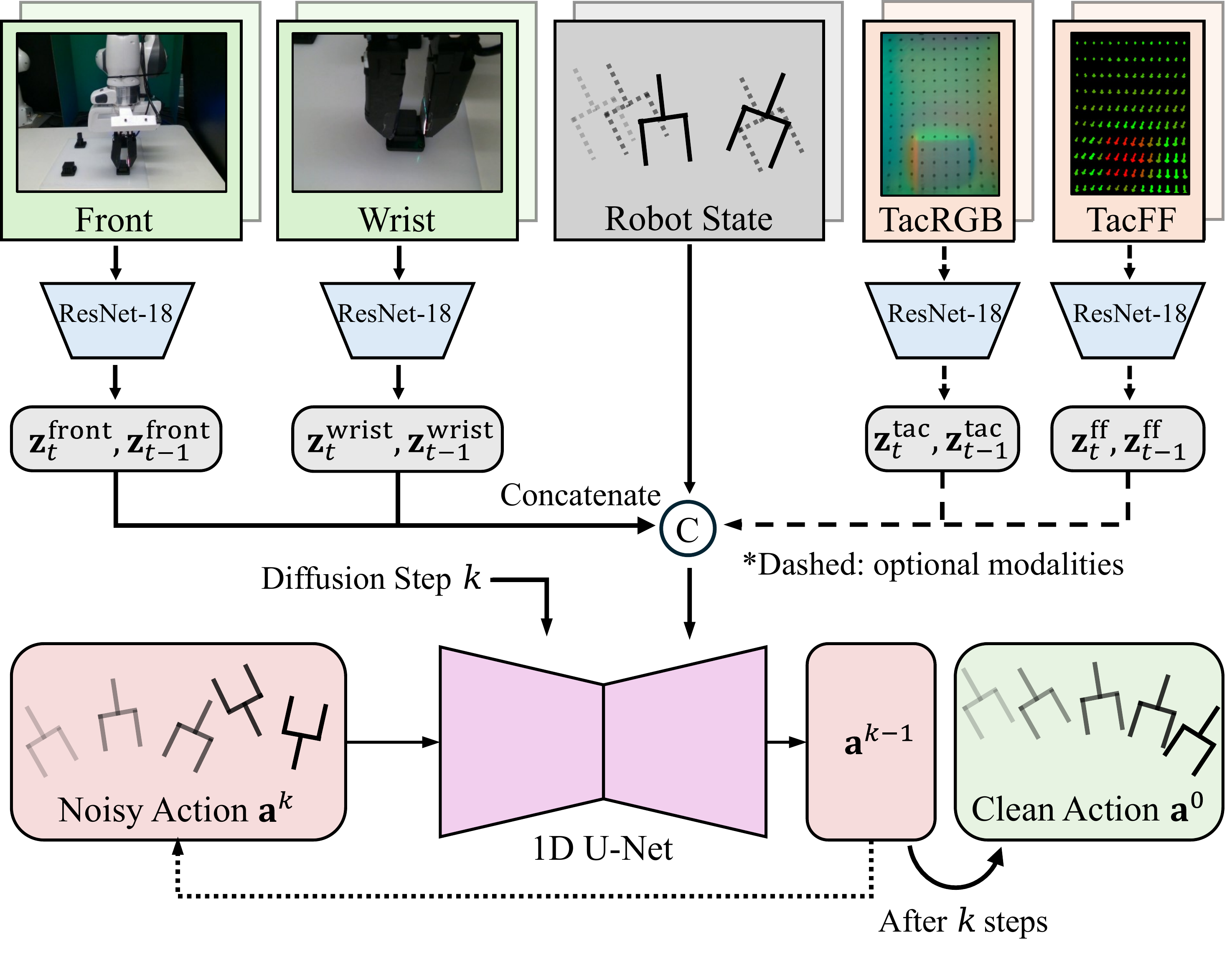}
    \caption{Overview of the visuotactile policy learning pipeline.
The policy conditions on a two-step observation history consisting of front- and wrist-view RGB images, the end-effector pose, and optional tactile inputs (TacRGB or TacFF; dashed). Encoded features are concatenated at node C (feature concatenation) and fed into a 1D U-Net–based diffusion model, which iteratively denoises action sequences to produce clean action chunks for execution.}
    \label{fig:policy_learning}
\end{figure}
We build upon the Diffusion Policy~\cite{diffusion_policy} framework and extend it to incorporate multimodal visuotactile observations. 
The resulting policy is trained via supervised imitation learning.

As illustrated in Fig.~\ref{fig:policy_learning}, the policy conditions on a two-step observation history 
$\mathcal{O}_t = \{O_{t-1}, O_t\}$.
At time $t$, the multimodal observation is defined as
\[
O_t = \{
I^{\mathrm{front}}_t,\ 
I^{\mathrm{wrist}}_t,\ 
I^{\mathrm{tac}}_t,\ 
I^{\mathrm{ff}}_t,\ 
\mathbf{p}_t
\},
\]
where $I^{\mathrm{front}}_t, I^{\mathrm{wrist}}_t \in \mathbb{R}^{3\times240\times320}$ 
denote RGB images captured by the front-view and wrist-mounted cameras, 
$I^{\mathrm{tac}}_t \in \mathbb{R}^{3\times240\times160}$ is the TacRGB image, 
$I^{\mathrm{ff}}_t \in \mathbb{R}^{3\times10\times14}$ represents the TacFF, 
and $\mathbf{p}_t \in \mathbb{R}^{7}$ denotes the end-effector state, consisting of 3D position and 4D quaternion orientation.

Each image-like modality is encoded using a ResNet-18 backbone, producing feature embeddings 
$\mathbf{z}^{\mathrm{front}}_t$, 
$\mathbf{z}^{\mathrm{wrist}}_t$, 
$\mathbf{z}^{\mathrm{tac}}_t$, and 
$\mathbf{z}^{\mathrm{ff}}_t$.
To incorporate temporal context, features from the two-step history 
$\{t-1, t\}$ are concatenated along the feature dimension.
The end-effector pose history is appended to form the fused conditioning embedding:
\begin{equation*}
\mathbf{z} = \mathrm{Concat}\Big(
\mathbf{z}^{\mathrm{front}}_{t-1:t},\ 
\mathbf{z}^{\mathrm{wrist}}_{t-1:t},\ 
\mathbf{z}^{\mathrm{tac}}_{t-1:t},\ 
\mathbf{z}^{\mathrm{ff}}_{t-1:t},\ 
\mathbf{p}_{t-1:t}
\Big).
\end{equation*}
Tactile features are included according to the selected sensing configuration.
For controlled comparison, we train separate policies under three settings:
(i) Vision Only,
(ii) Vision + TacRGB, and
(iii) Vision + TacFF.

Since the disassembly tasks primarily involve vertical pulling motions, the action space is restricted to 3D translation and yaw rotation. 
Each action is defined as a 5D relative command consisting of position increments  $(\Delta x, \Delta y, \Delta z)$,  a rotation increment about the world-frame $z$-axis $\Delta \theta_z$, and a gripper command $g$.

Given the conditioning embedding $\mathbf{z}$,  a 1D U-Net parameterizes the denoising network of the diffusion model to generate an action sequence  $\mathbf{a}_{t:t+H-1} \in \mathbb{R}^{H\times 5}$ with horizon $H = 16$. 
At inference time, DDIM sampling with $K = 10$ denoising steps produces an action chunk,  of which only the first $8$ actions are executed before replanning.
\section{EXPERIMENTS}
\subsection{Experimental Setup}
We evaluate our approach in both simulation and real-world settings under a unified sensing and policy framework. 
At each control step, the policy receives multimodal observations consisting of front and wrist RGB images, tactile sensing (TacRGB and TacFF), and the end-effector state. 
All RGB images are resized to $240 \times 320$, while TacRGB is cropped to $240 \times 160$. 
TacFF is represented as a $10 \times 14$ grid, where each cell encodes shear-$x$, shear-$y$, and normal components.
All modalities are synchronized at 10~Hz and normalized to $[-1,1]$ prior to feature encoding.
\subsubsection{Simulation Environment}
All simulated tasks are implemented in IsaacGym~\cite{makoviychuk2021isaac} with TacSL~\cite{akinola2025tacsl} for vision-based tactile rendering. 
The simulated agent mirrors the real-world sensing setup, including front and wrist cameras and an end-effector tactile sensor.

During evaluation, object and socket positions are randomized within $\pm 5$\,cm in the $x$–$y$ plane, and in-hand orientations are perturbed by up to $10^{\circ}$~\cite{luu2025manifeel}. 
A rollout is deemed successful if the target component is fully extracted and lifted above a predefined height threshold.
Success rates are averaged over the final 10 training epochs across three random seeds. 
For each seed and epoch, 50 distinct environment initializations are evaluated, yielding 1500 rollouts per task.

\subsubsection{Real-World}
\begin{figure}[t]
    \centerline{\includegraphics[width=1.0\linewidth]{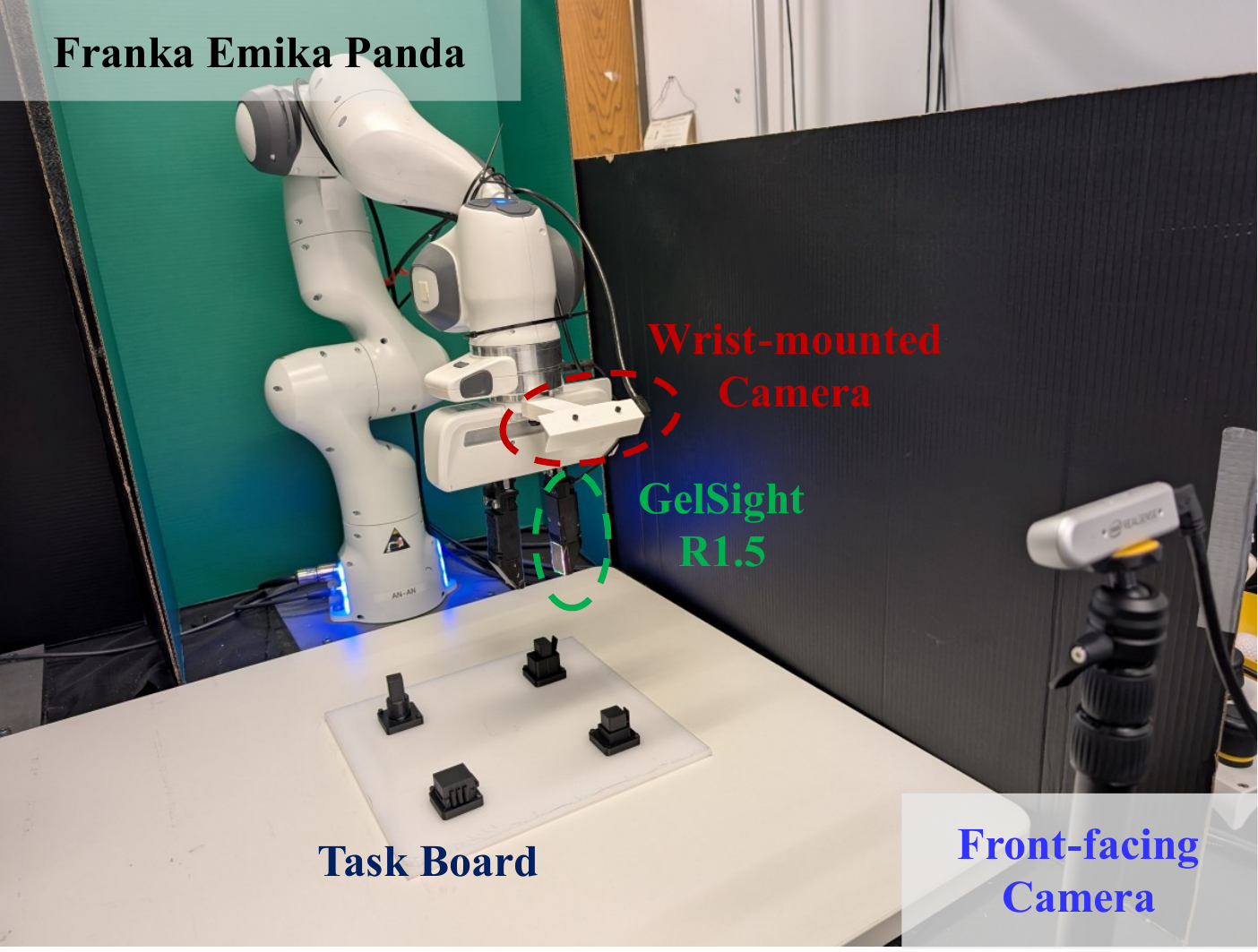}}
    \caption{Real-world experimental setup, including the Franka Emika Panda manipulator, a wrist-mounted RealSense D415 camera, and a front-facing RealSense D415 camera.}
    \label{fig:hardware_configuration}
\end{figure}

The real-world setup is shown in Fig.~\ref{fig:hardware_configuration}. 
Experiments are conducted using a Franka Emika Panda manipulator equipped with a wrist-mounted RealSense D415 camera and a front-facing RealSense D415 camera. 
Tactile feedback is provided by a GelSight R1.5~\cite{wang2021gelsight} sensor mounted on the right finger.

For each task, 50 demonstrations with randomized
object poses are collected via teleoperation using a 3Dconnexion SpaceMouse. 
The recorded multimodal observations are used to train diffusion policies for 400 epochs under each modality configuration. 
During rollout, we apply DDIM~\cite{DDIM} sampling with 10 denoising steps for efficient inference.

\subsection{Simulation Results}
\begin{table*}[t]
    \centering
    \caption{Success rates across disassembly tasks and sensing modalities in simulation.}
    \label{tab:result_sim}
    \small
    \newcolumntype{Y}{>{\centering\arraybackslash}X}
    \begin{tabularx}{\textwidth}{l|Y Y Y Y Y}
        \toprule
         & \textbf{Task S1} & \textbf{Task S2} & \textbf{Task S3} & \textbf{Task S4} & \textbf{Task S5} \\
        \textbf{Modality} & Vertical Pull, & Vertical Pull, & Loose Plug & Vertical Pull, & Vertical Pull \\
         & Loose Socket & Tight Socket & with Lid & Flat Barb & Spike Barb \\
        \midrule
        \textbf{Vision Only} & 81.7\% & 36.9\% & 50.0\% & 74.6\% & 59.5\% \\
        \addlinespace
        \textbf{Vision + TacRGB} & 84.4\% & 37.2\% & 51.4\% & 75.1\% & 57.8\% \\
        \addlinespace
        \textbf{Vision + TacFF} & \textbf{86.5\%} & \textbf{44.3\%} & \textbf{53.5\%} & \textbf{78.8\%} & \textbf{61.9\%} \\
        \bottomrule
    \end{tabularx}
    
\vspace{0.3em}
\footnotesize
\parbox{1.0\linewidth}{
We report the mean success rate, computed as the average over the final 10 checkpoints, each evaluated in 50 environment initializations. Results are further averaged over 3 training seeds, yielding 1500 evaluations per policy. \textbf{Bold} indicates the best performance.}
\end{table*}

\begin{table*}[t]
    \centering
    \caption{Success rates across disassembly tasks and sensing modalities in real-world experiments.}
    \label{tab:result_real_normal_rate}
    \small
    \newcolumntype{Y}{>{\centering\arraybackslash}X}
    \begin{tabularx}{\textwidth}{l|Y Y Y Y Y}
        \toprule
         & \textbf{Task R1} & \textbf{Task R2} & \textbf{Task R3} & \textbf{Task R4} & \textbf{Task R5} \\
        \textbf{Modality} & Vertical Pull, & Vertical Pull, & Loose Plug & Vertical & Vertical \\
         & Loose Socket & Tight Socket & with Lid & Push Tab & Clip \\
        \midrule
        \textbf{Vision Only} & 80.0\% & 55.0\% & 40.0\% & 70.0\% & 15.0\% \\
        \addlinespace
        \textbf{Vision + TacRGB} & 90.0\% & 30.0\% & 15.0\% & 5.0\% & 45.0\% \\
        \addlinespace
        \textbf{Vision + TacFF} & \textbf{95.0\%} & \textbf{70.0\%} & \textbf{45.0\%} & \textbf{75.0\%} & \textbf{55.0\%} \\
        \bottomrule
    \end{tabularx}
    
\vspace{0.3em}
\footnotesize
\parbox{1.0\linewidth}{
Each real-world policy was evaluated over 20 independent trials.}
\end{table*}

\begin{table*}[t]
    \centering
    \caption{Average execution time computed over successful disassembly rollouts in real-world experiments.}
    \label{tab:result_real_normal_time}
    \small
    \newcolumntype{Y}{>{\centering\arraybackslash}X}
    \begin{tabularx}{\textwidth}{l|Y Y Y Y Y}
        \toprule
         & \textbf{Task R1} & \textbf{Task R2} & \textbf{Task R3} & \textbf{Task R4} & \textbf{Task R5} \\
        \textbf{Modality} & Vertical Pull, & Vertical Pull, & Loose Plug & Vertical & Vertical \\
         & Loose Socket & Tight Socket & with Lid & Push Tab & Clip \\
        \midrule
        \textbf{Vision Only} & \textbf{21.1 s} & 25.4 s & 51.4 s & 19.9 s & \textbf{28.3 s} \\
        \addlinespace
        \textbf{Vision + TacRGB} & 22.4 s & 30.2 s & \textbf{28.0 s} & 19.0 s & 35.9 s \\
        \addlinespace
        \textbf{Vision + TacFF} & 22.8 s & \textbf{22.5 s} & 67.6 s & \textbf{18.3 s} & 58.0 s \\
        \bottomrule
    \end{tabularx}
\end{table*}

Table~\ref{tab:result_sim} reports the success rates across all simulated disassembly tasks. 
Since each configuration is evaluated over 1500 rollouts, even modest percentage-point differences reflect statistically meaningful performance trends.

\subsubsection{Overall Performance and Geometric Constraints}
Across the five simulated tasks, incorporating TacFF yields the highest success rates. 
The performance gap between modalities becomes more pronounced as geometric constraints tighten. 
In Task S1, generous clearance results in a high baseline success rate, and tactile inputs provide marginal gains. 
In contrast, Task S2 imposes stricter alignment and friction constraints. 
Here, the Vision Only policy drops to 36.9\%, whereas integrating TacFF improves performance to 44.3\%. 
These results suggest that under tighter geometric tolerances, the low-dimensional normal and shear force cues encoded in TacFF help resolve contact ambiguities that are difficult to infer from vision alone.

\subsubsection{Asymmetric Resistance and Interlocking Constraints}

Tasks S4 and S5 introduce asymmetric resistance through flat and spike-shaped barb structures. 
In these scenarios, direct vertical pulling leads to failure; successful execution requires first sliding to disengage the barb, followed by controlled upward extraction.
In Task S4, TacFF effectively captures shear patterns during the sliding phase, increasing the success rate to 78.8\%. 
In Task S5, the Vision + TacRGB configuration slightly underperforms the Vision Only baseline (57.8\% vs.\ 59.5\%), whereas TacFF improves performance to 61.9\%. 
This contrast highlights a representation mismatch. 
TacRGB provides high-resolution surface deformation signals, which may introduce task-irrelevant variability when the key requirement is identifying a directional escape motion. 
In comparison, TacFF directly encodes distributed shear forces, offering compact and structured cues that facilitate coordinated sliding and pulling without jamming.

Overall, the simulation results indicate that in rigid-body disassembly tasks characterized by tight tolerances and interlocking constraints, structured force representations such as TacFF provide more effective guidance than high-dimensional TacRGB.

\subsection{Real-World Results}
We next evaluate policy performance on physical hardware. 
Each task (R1–R5) is executed for 20 rollouts under standard indoor lighting condition, with object poses randomized within the same range used during data collection.

\subsubsection{Success Rate}

Table~\ref{tab:result_real_normal_rate} reports success rates across all real-world tasks. 
The trends observed in simulation largely carry over to the physical setting.
For relatively simple tasks (R1, R2), performance remains high across configurations. 
In more contact-sensitive scenarios, particularly R4 and R5, differences between sensing modalities become more pronounced. 
TacFF configuration consistently achieves the strongest performance, including 75.0\% success in R4 despite the presence of deformable components.
Tasks R3 and R5 exhibit lower overall success rates (45.0\% and 55.0\%), reflecting their tighter alignment requirements and more complex contact interactions. 
These results suggest that real-world contact uncertainty and manipulation precision significantly influence performance.

Across tasks, structured force feedback provides clearer guidance during extraction compared to purely visual input, especially when visual cues are ambiguous or occluded. 
In contrast, adding TacRGB does not consistently yield further improvements, indicating that increasing observation dimensionality alone does not guarantee better performance.

\subsubsection{Execution Time}

Table~\ref{tab:result_real_normal_time} reports the average completion time over successful rollouts in real-world experiments. In several tasks, policies equipped with TacFF exhibit longer execution times.
This behavior is expected for two reasons. First, incorporating an additional tactile modality increases the computational and perceptual processing involved in decision making. More importantly, TacFF provides fine-grained force feedback, enabling the policy to actively adjust and explore contact states when the initial attempt is unsuccessful. As a result, the policy may perform multiple corrective adjustments before achieving successful disassembly.
In contrast, vision-only and Vision + TacRGB policies rely primarily on visual cues and lack force information for fine contact regulation. When an initial manipulation attempt fails, these policies are more likely to terminate unsuccessfully rather than continue corrective exploration. Since the reported time is averaged only over successful rollouts, TacFF policies exhibit longer execution times in cases where additional exploratory adjustments are required.

\subsection{Robustness and Modality Analysis}
\subsubsection{Dim Lighting Condition}
To evaluate robustness under degraded visual perception, we conduct additional experiments under dim lighting for Tasks R1 and R5. 
For this setting, 50 demonstrations are collected with randomized object poses, and separate policies are trained for each sensing configuration using 400 epochs.

As shown in Table~\ref{tab:result_real_dim}, the effect of reduced illumination varies across tasks. 
In Task R1, all configurations exhibit a noticeable decrease in success rate. 
In contrast, for Task R5, the Vision Only policy degrades sharply (from 15.0\% to 0.0\%), whereas Vision + TacRGB and Vision + TacFF remain comparatively stable. 
Notably, Vision + TacFF maintains identical performance (55.0\%) under both normal and dim lighting.
These results indicate that the contribution of tactile sensing is task-dependent. 
Task R1 appears primarily governed by geometric alignment cues, making it sensitive to visual degradation. 
In contrast, Task R5 involves contact-driven interactions where tactile feedback provides more reliable information. 
Under such conditions, tactile-enhanced configurations exhibit improved robustness when visual signals become unreliable.

Overall, tactile sensing serves not merely as a supplementary modality but as a stabilizing signal in tasks dominated by contact dynamics.

\begin{table}[t]
    \centering
    \caption{Success rates across tasks and modalities in real-world experiments under normal and dim lighting conditions.}
    \begin{tabular*}{\columnwidth}{@{\extracolsep{\fill}}l|cccc}
        \toprule
         & \multicolumn{2}{c}{\textbf{Task R1}} & \multicolumn{2}{c}{\textbf{Task R5}}  \\
        \textbf{Modality} & \multicolumn{2}{c}{Vertical Pull,} & \multicolumn{2}{c}{\multirow{2}{*}{Vertical Clip}} \\
         & \multicolumn{2}{c}{loose socket} &  \\
        \cmidrule(lr){2-3} \cmidrule(lr){4-5}
        Lighting & Normal & Dim & Normal & Dim \\
        \midrule
        \textbf{Vision Only} & 80.0\% & 25.0\% & 15.0\% & 0.0\% \\
        \addlinespace
        \textbf{Vision + TacRGB} & 90.0\% & 10.0\% & 45.0\% & 40.0\% \\
        \addlinespace
        \textbf{Vision + TacFF} & \textbf{95.0\%} & \textbf{25.0\%} & \textbf{55.0\%} & \textbf{55.0\%} \\
        \bottomrule
    \end{tabular*}
    \label{tab:result_real_dim}
\end{table}

\subsubsection{Modality Variations}
To examine whether combining tactile representations yields further gains, we evaluate an additional configuration that incorporates Vision, TacRGB, and TacFF. 
Experiments are conducted on Tasks R3 and R5 under normal lighting, using the same demonstration datasets and training protocol as the previous configurations.

As shown in Table~\ref{tab:result_real_all}, the combined configuration does not improve performance. 
Success rates decrease to 0.0\% for Task R3 and 20.0\% for Task R5, both substantially lower than those achieved by Vision + TacFF.
This result suggests that naive concatenation of heterogeneous sensory inputs does not necessarily lead to improved performance. Simply increasing the dimensionality of the observation space may prevent the policy from focusing on task-relevant signals.

These findings indicate that effective multimodal fusion requires more than straightforward feature aggregation. 
More structured fusion strategies, such as modality-aware attention mechanisms or dedicated encoders tailored to each sensing stream, may be necessary to properly integrate heterogeneous tactile representations without diluting task-relevant signals.

\begin{table}[t]
    \centering
    \caption{Effect of combining TacRGB and TacFF on real-world success rates.}
    \begin{tabular*}{\columnwidth}{@{\extracolsep{\fill}}l|cc}
        \toprule
         & \textbf{Task R3} & \textbf{Task R5}  \\
        \textbf{Modality} & Loose Plug with Lid & Vertical Clip \\
        \midrule
        \textbf{Vision Only} & 40.0\% & 15.0\% \\
        \addlinespace
        \textbf{Vision + TacRGB} & 15.0\% & 45.0\% \\
        \addlinespace
        \textbf{Vision + TacFF} & \textbf{45.0\%} & \textbf{55.0\%} \\
        \addlinespace
        \textbf{Vision + TacRGB + TacFF} & 0.0\% & 20.0\% \\
        \bottomrule
    \end{tabular*}
    \label{tab:result_real_all}
\end{table}

\section{CONCLUSIONS} 
In this work, we systematically investigated the role of tactile sensing in robotic disassembly through a structured evaluation spanning simulation and real-world settings. 
Across tasks with increasing geometric constraints, asymmetric resistance, and deformable interactions, force-field-based tactile representations consistently improved robustness compared to vision-only policies.

Our results reveal a clear task-dependent pattern. 
In geometry-dominant scenarios with sufficient visual cues, tactile sensing provides marginal gains. 
However, as contact uncertainty increases or visual information becomes unreliable, structured force representations play a critical role in resolving contact ambiguities and stabilizing manipulation.
Importantly, we show that simply increasing observation dimensionality does not guarantee improved performance. 
Naively combining heterogeneous tactile representations can degrade performance, highlighting the need for more principled multimodal integration strategies.

Overall, our findings suggest that in contact-rich disassembly, compact and structured force encoding is more effective than high-dimensional tactile imagery, and that effective multimodal learning depends not only on adding modalities but on how they are represented and fused.

\bibliographystyle{unsrt}
\bibliography{references}

\end{document}